\pdfoutput=1

\documentclass[11pt]{article}

\usepackage{acl}

\usepackage{multirow}
\usepackage{booktabs}
\usepackage{times}
\usepackage{latexsym}
\usepackage{graphicx}
\usepackage{arydshln}

\usepackage[T1]{fontenc}

\usepackage[utf8]{inputenc}

\usepackage{microtype}

\newenvironment{enumerate*}%
  {\begin{enumerate}%
  
    \setlength{\itemsep}{.2pt}%
    \setlength{\parskip}{.2pt}%
    \setlength{\topsep}{.5pt}}%
  {\end{enumerate}}

\newenvironment{itemize*}%
  {\begin{itemize}%
    \setlength{\itemsep}{.2pt}%
    \setlength{\parskip}{.2pt}%
    \setlength{\topsep}{.5pt}}%
  {\end{itemize}}

\title{Reasoning Circuits: Few-shot Multihop Question Generation with Structured Rationales}

\author{
Saurabh Kulshreshtha and Anna Rumshisky\\
Department of Computer Science\\
University of Massachusetts Lowell\\
\texttt{\{skul,arum\}@cs.uml.edu}\\
}

\begin{document}
\maketitle
\begin{abstract}
Multi-hop Question Generation is the task of generating questions which require the reader to reason over and combine information spread across multiple passages using several reasoning steps. Chain-of-thought rationale generation has been shown to improve performance on multi-step reasoning tasks and make model predictions more interpretable. However, few-shot performance gains from including rationales have been largely observed only in +100B language models, and otherwise require large-scale manual rationale annotation. In this paper, we introduce a new framework for applying chain-of-thought inspired structured rationale generation to multi-hop question generation under a very low supervision regime (8- to 128-shot). We propose to annotate a small number of examples following our proposed multi-step rationale schema, treating each reasoning step as a separate task to be performed by a generative language model. We show that our framework leads to improved control over the difficulty of the generated questions and better performance compared to baselines trained without rationales, both on automatic evaluation metrics and in human evaluation.  Importantly, we show that this is achievable with a modest model size. 
\end{abstract}


\section{Introduction}
Recently, there has been a surge of interest in the NLP community in 
%
the idea of providing supervision to language models(LMs) in the form of human-written rationales \cite{wiegreffe-marasovic-2021-review, NEURIPS2018_4c7a167b, jansen-etal-2018-worldtree, aggarwal-etal-2021-explanations, geva-etal-2021-aristotle, inoue-etal-2020-r4c} which explain why and how the target label is arrived at. Using human-written explanations as an intermediate step has been shown to improve performance on a variety of predictive tasks, compared to the cases where no rationales are provided \cite{wiegreffe-marasovic-2021-review}. However, rationales are expensive to collect through manual annotation at a large scale. 

Explanations can take several forms, such as textual highlights, free-text explanations and structured explanations. In this work, we focus on the latter two. By rationales we refer to several structured sentences enumerating intermediate steps of reasoning required to solve a multi-step reasoning problem before producing the target text.

In chain-of-thought rationale generation paradigm \cite{cot,star}, LMs learn to generate rationales - a step toward explainable NLP models. Prompting LMs with few-shot rationale examples has been shown to improve performance for multi-step reasoning tasks compared to standard prompting without rationales \cite{cot}. However, this effect is only observed in extremely large language models (XLLMs) with +100b parameters \cite{cot, dmcot}. At the same time access to XLLMs is limited in the research community due to  costs and infrastructure required to fine-tune and inference them. In many cases these models are never released publicly.

In essence, supervision from rationales is a richer signal compared to supervision from only target labels, especially for multi-step reasoning tasks. However, only XLLMs have been shown to capture this signal in few-shot regimes. In this paper we assume large-scale rationale annotation to remain unavailable since it is tedious and arguably more expensive to generate than standard target label annotation. This leads to a currently unaddressed challenge of solving multi-step problems by smaller LMs than XLLMs in a few-shot regime.

This work deals with the complex task of Multi-hop Question Generation(MQG) where, given multiple passages and a pre-defined answer, the objective is to generate challenging questions that cannot be answered only from reading a single passage, this task requires many steps of reasoning to accomplish and we further constrain ourselves to the case where supervision available is restricted to a few number of labelled examples.

Our contributions can be summarized as follows. We propose a new framework called Reasoning Circuits applicable specifically for the often encountered constraints faced in real-world where:

\noindent1. Large-scale annotation is not possible or available, only a limited number of examples of a multi-step reasoning problem are available.

\noindent2. Access limited to modest neural compute infrastructure that can support training models up to a maximum of 3 billion parameters.

\noindent3. Budget for rationale annotation is limited.

In this work we apply this framework to MQG task in a few-shot setting. This entails identifying reasoning steps human annotators employ to generate multi-hop questions and codifying them into a structured rationale annotation scheme, and manually producing rationale annotations for the few examples, capped at a maximum of around 200 examples. A generative model is then fine-tuned with a mixture of tasks where each "task" refers to a single smaller step of reasoning derived from the structured rationales designed for the MQG task. We report improvements over baselines where no rationale was employed on automatic evaluation metrics as well as human evaluation. We also show reduced gap in performance between our system only trained with approimately 150 examples(training and validation combined) and prior art that has been trained with 9,000 to 90,000 examples without rationales on automatic evaluation metrics.

\section{Related Work}
\subsection{Multi-hop Question Generation}
Several research studies focus on the task of single-hop question generation on datasets like SQuAD \cite{rajpurkar-etal-2016-squad} for instance, \citet{kim2019improving} propose ASs2s-a, a seq2seq model based on Long Short-term Memory (LSTM), which separately encodes answer and context.

There are studies about generating more difficult questions on knowledge graphs which includes \citet{talmor-berant-2018-web} and \citet{kumar}. However, these are not directly applicable to free-text since, it is not made up of entity relation triplets, as is the case with knowledge bases.

Proposed systems for MQG with free-text, SGGDQ-DP \cite{pan-etal-2020-semantic}, MultiQG \cite{su-etal-2020-multi}, DFGN+QG \cite{yu1} and GATENLL+CT \citet{sachan} rely on external tools like name entity recognition, entity linking and coreference resolution to construct knowledge graphs with which complex questions are generated with decoders. Closely related to our work \citet{cheng-etal-2021-guiding} propose to control question difficulty, by progressively increasing question hops through step-by-step rewriting with GPT2-small\cite{gpt2} under the guidance of an extracted reasoning chain, generated also from external tools. QA4QG \cite{qa4qg} is current state-of-the art for MQG task, where attention patterns of a multi-hop question answering model  guide a MQG model.

In the F+R+A system proposed by \citet{xie-etal-2020-exploring} reinforcement rewards for fluency, relevance and particularly answerability - also generated by a separate QA model, are introduced in tandem with standard cross-entropy loss for MQG. In SemQG \cite{zhang-bansal-2019-addressing}  two semantics-enhanced rewards are proposed to regularize a question generation model. ADDQG \cite{wang-etal-2020-answer} treats semantic and syntactic metrics as reinforcement rewards for MQG task.

All systems cited until now utilise large scale supervision of 90k training examples from HotpotQA dataset \cite{yang-etal-2018-hotpotqa} with the exception of \citet{cheng-etal-2021-guiding} at 57k. LowResourceQG system \cite{yu-etal-2020-low} learns the structural patterns from unlabeled questions and transfers this to a MQG model, it is train with 9,000 examples from HotpotQA dataset. 

\subsection{Few-shot Rationale Generation}
XLLMs can learn to generate valid rationales for multi-step problems with few-shot in-context learning examples \cite{cot, dmcot}, smaller models on the other hand, need to be trained on more rationale annotation to achieve strong performance. To reduce the dependence on tedious rationale annotations a hybrid self-learning approach with smaller 6B parameter LMs has been proposed by \citet{star} that uses fewer rationale annotations, however even this method requires a lot of manually annotated data. Since the generated silver rationales are noisy, to filter and improve these rationales large-scale standard input-target annotation is required. The ground truth references are used as a proxy to check and filter out generated silver rationales, by comparing the references to the predicted target text, produced along with the rationales. Ground truth references also are used to provide hints to the model when it fails to generate the correct answer. Filtered silver rationales thus accumulated are then used for iterative self-training until performance plateaus.

\section{Structured Rationales for Multi-hop Question Generation}
HotpotQA dataset \cite{yang-etal-2018-hotpotqa} is one of the most widely used benchmarks of multi-hop question answering and consist of broadly two types of questions: bridge-entity questions and comparison questions.  In bridge-entity questions, which constitute 75\% of the dataset, annotators get pairs of passages where at least one entity (called the bridge entity) is present in both passages. In comparison questions, annotators are provided a pair of passages about entities drawn from a similar theme such as musicians, authors, films, plants among several categories. A comparison question typically compares some quality of the central entities in the two passages. Question types found in this dataset cover 5 out of 6 total sub-types of multi-hop questions as identified in recent survey on multi-hop question answering and generation \cite{surveymulti} originally identified in \citet{min-etal-2019-multi} with the only exception of commonsense reasoning.

Below, we identified the reasoning steps that annotators needed to follow in order to create the questions of each type.

\subsection{Bridge Entity Questions}
\begin{figure*}[ht]
    \centering
    \includegraphics[width=1.0\linewidth]{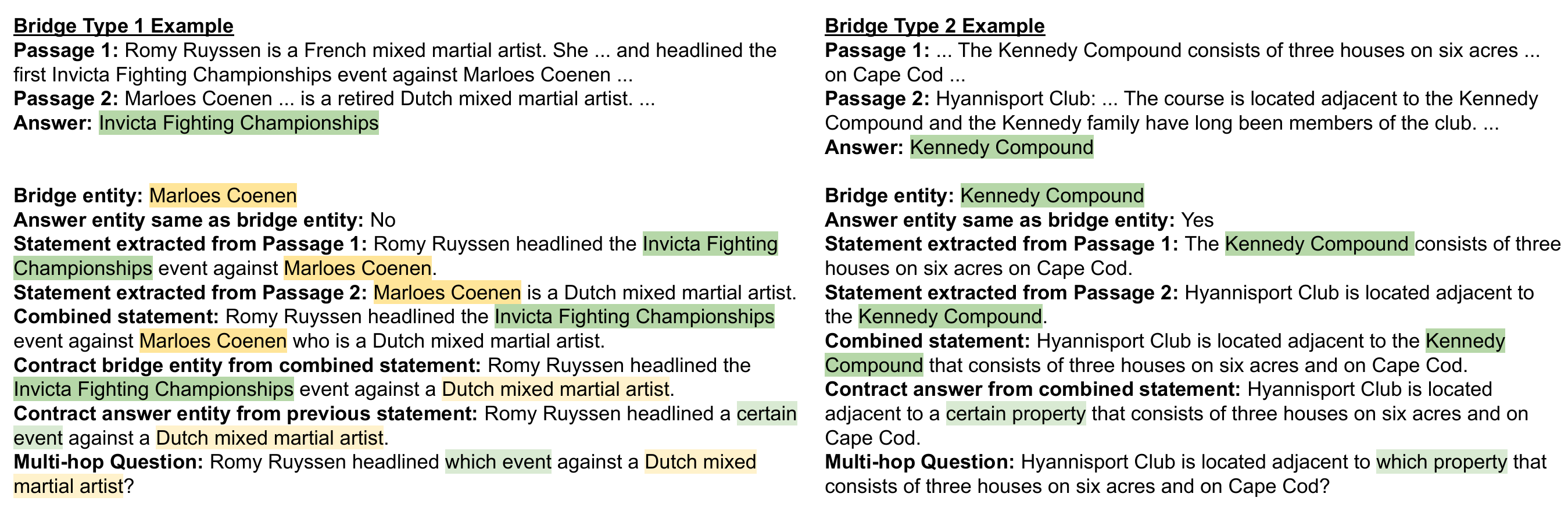}
\vspace{-0.5cm}
    \caption{Two examples of bridge rationales that lead to the creation of multi-hop question. Example 1 shows rationale annotation when answer span is not found in both passages. Example 2 shows another of example of rationale annotation when the answer is present in both passages, so that the Step 4 is skipped. The highlights in green and yellow show the answer and bridge entities, in the second example they are the same. In the penultimate steps, the lighter highlights indicate the substitution of bridge and answer entities with a common noun preceded by "certain". In the last step typically a Wh- word substitutes "certain" or "some" words, however more radical transformations also take place in our annotations.}
    \label{fig:bridge}
\end{figure*}

 Figure \ref{fig:bridge} shows examples of reasoning steps for bridge-type multihop questions. Given two passages and a pre-defined answer to the question to be generated, an annotator would need to:
\begin{enumerate*}
\item Select the bridge entity $b$, an entity present in both passages. If the answer is present in both passages, bridge entity is set to the answer.
\item From each passage ($p_1$ and $p_2$), extract one statement ($s_1$ and $s_2$) about the bridge entity which connects the answer to the bridge entity, if the passage contains the answer.
\item Combine the two statements ($s_1$ and $s_2$) into a single combined statement $c$.
\item Substitute bridge entity $b$ in combined statement $c$ with a common noun to get $c-b$. For example, replace "the Beatles" with "a band".
\item Substitute the answer $a$ in $c-b$ with a common noun preceded by "certain" or "some" to get $c-b-a$. For example, the answer entity "5th March 1992" is replaced by "certain date" or "someday" and the answer "George Orwell" is replaced by "certain person" or "someone".
\item Convert the statement from the previous step into a question, such that the answer to it is the provided answer span.
\end{enumerate*}
\noindent
 Step 4 is skipped if bridge and answer are the same.

\subsection{Comparison Questions}
\begin{figure*}[ht]
    \centering
    \includegraphics[width=1.0\linewidth]{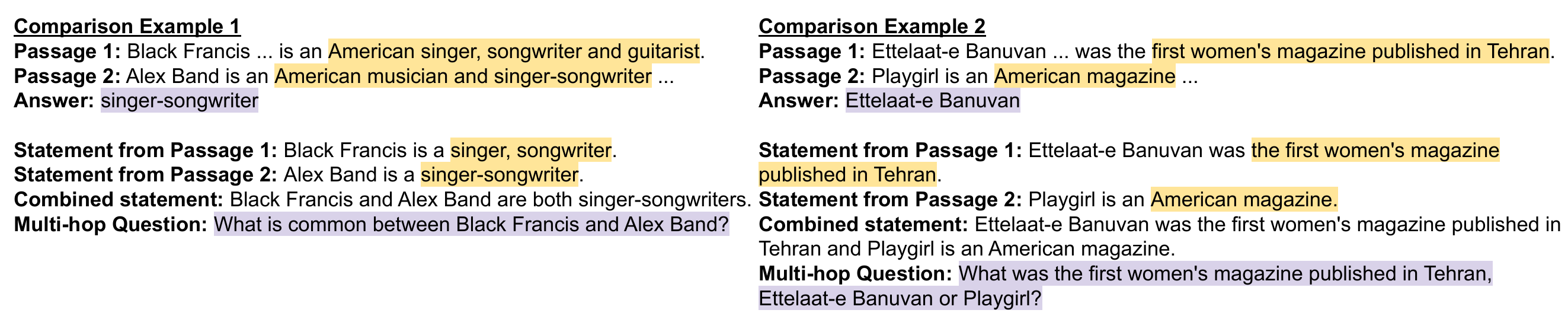}
\vspace{-0.5cm}
    \caption{Two examples of comparison multi-hop question. In Example 1 similar attributes of the central figures in the two passages have been highlighted, later this is turned into a similarity multi-hop question. In Example 2 difference between the two magazines is used as a basis to create a multi-hop question.}
    \label{fig:comparison}
\end{figure*}

Figure \ref{fig:comparison} shows two examples of this type. For this question type, given two passages and a pre-defined answer, an annotator would need to:
\begin{enumerate*}
\item Extract two statements($s_1$ and $s_2$), one from each passage, such that a comparison can be drawn between the two statements, keeping the answer in mind.
\item The two statements are combined into a single statement($c$), highlight the nature of similarity or difference between information from the two paragraphs, keeping the answer in mind.
\item Generate a comparative question($q_c$) from reading the combined statement and answer.
\end{enumerate*}

\section{Rationale Steps as a Mixture of Tasks}
The step-by-step nature and clearly defined structure of the rationales identified above motivated us to formulate this multi-step problem into a mixture of tasks. A step of reasoning is treated as a task, only the information required to perform the reasoning step is treated as the input and the result of the reasoning step is the expected output.

We formalise bridge multi-hop question generation into nine separate smaller steps of reasoning and we identify 3 steps of reasoning for generating comparison multi-hop questions. Additionally, a reasoning step should identify whether a given pair of passages and answer are more suitable to generate a bridge question from or a comparison question instead. For this purpose we create the first task of predicting the question type.

These reasoning steps or tasks can be categorized into: 1. Control tasks where the outcome is a control variable whose value decides what reasoning path to follow and 2. Generative tasks which essentially generate free-form text to be treated as input for later reasoning steps or as the final output. We describe each of these tasks below.

\subsection{Question Type Task}
\textbf{Task 1} is a control task that decides the question type. Given input passages $p1$, $p2$ and the answer $a$, this task assigns the control variable $q_{type}$ either the value $bridge$, $comparison$ or $confused$. The values $bridge$, $comparison$ are assigned if the model literally generates the tokens "bridge" and "comparison" which lead the current input example to only follow either the bridge or the comparison tasks trajectories. In case the model gets confused and fails to generate "bridge" or "comparison" as its output during test time we assign $q_{type}$ the value $confused$ which leads to both bridge and comparison task trajectories to be followed and two questions  one of each type are generated.

\subsection{Bridge Rationale Steps}

\textbf{Task 2} generates the bridge entity $b$ provided input passages $p_1$, $p_2$ and the answer $a$.

\noindent\textbf{Task 3} is a control task that identifies whether the answer span($a$) is present in the provided passage($p_i$) or not and assigns the boolean value of $True$ or $False$ to the variables $in^{a}_{p_i}$ for $i$ in $\{1,2\}$ depending on whether whether the model generates the token "present" or "absent".

\noindent\textbf{Task 4} is a control task that identifies whether the answer span($a$) is the same as the bridge entity ($b$) or not and assigns a value of either $True$ or $False$ to the variable $same^{a}_{b}$. 

\noindent\textbf{Task 5} generates a statement $s_{i}$ which connects the bridge entity $b$ to the answer entity $a$ from the passage $p_i$. This task is only run on a passage that contains the answer entity or $in^{a}_{p_i} = True$ unless the answer entity is the same as the bridge entity and present in both passages in which case this task is not run on either of the passages.

\noindent\textbf{Task 6} generates a statement $s_{i}$ about bridge entity ($b$) from the passage $p_i$. Inputs include the $i^{th}$ input passage $p_i$ and the bridge entity $b$. When the $same^{a}_{b} = True$, Task 6 is run on both $p_1$ and $p_2$ to get $s_{1}$ and $s_{2}$. Otherwise it is only run on the passage $p_i$ where $in^{a}_{p_i} = False$

\noindent\textbf{Task 7} generates a single combined statement $c$ from statements $s_{1}$ and $s_{2}$. Inputs to this step include the answer $a$ the bridge entity $b$ and the generated statements $s_{1}$ and $s_{2}$ from prior steps.

\noindent\textbf{Task 8} This task contracts the bridge entity from the combined sentence $c$ and substitutes it with a common noun to get $c-b$. It is a generative task and inputs to this step include combined statement ($c$), the answer ($a$) and the bridge entity ($b$).

\noindent\textbf{Task 9} contracts the answer entity in the sentence $c-b$ and substitutes it with a indefinite determiner followed by a common noun to get $c-b-a$. It is a generative task and inputs to this step include combined statement with bridge contracted ($c-b$), the answer ($a$) and the bridge entity ($b$).

\noindent\textbf{Task 10} transforms the combined statement with both the answer and bridge contracted ($c-b-a$) into a multi-hop question $q_b$ enquiring about the noun which is preceded by "certain" or the some-word. It is the final reasoning step in the bridge type rationales and is a generative task. Inputs to this step include $c-b-a$, $c-b$ and $a$.

\subsection{Comparison Rationale Tasks}
\textbf{Task 11} simultaneously generates both statements $s_{1}$ and $s_2$ which deliberate over a similar or dissimilar quality about the key entities in passages $p_1$ and $p_2$ respectively. Input includes passages $p_1$, $p_2$ and the answer $a$. The reason for concurrently producing both $s_{1}$ and $s_2$ is to maintain the same decoder state while producing both $s_1$ and $s_2$.

\noindent\textbf{Task 12} generates a single combined statement $c$, a conjunction of the two statements $s_{1}$, $s_{2}$ emphasis is on comparison between the two. It is a generative task, and serves as the second reasoning step in comparison rationales. Inputs to this step include the generated statements $s_{1}$, $s_{2}$ and answer $a$.

\noindent\textbf{Task 13} transforms the combined statement $c$ into a comparitive question $q_c$. It serves as the third and final reasoning step in comparison rationales. Inputs to this step include the combined statement $c$ and the answer $a$ as inputs.

\begin{figure*}[!htb]
    \centering
    \includegraphics[width=1.0\linewidth]{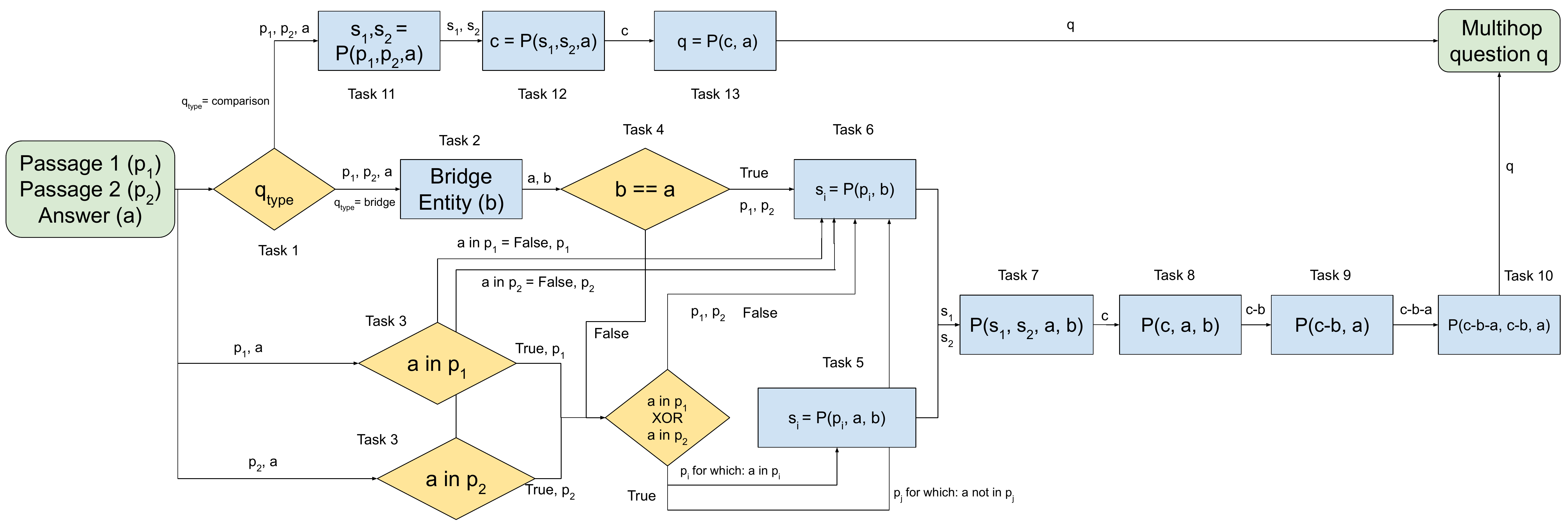}
\vspace{-0.5cm}
    \caption{Reasoning Circuits: This flowchart depicts the connections between different reasoning steps/tasks. Control tasks are marked as yellow decision nodes while the generative tasks are marked by blue process nodes. Edge labels denote the significant inputs that flow from one task to next as well as the value of the control signal at the root of the edge that enables this edge flow data flow to be enabled. Note that a single LM is fine-tuned on a mixture of all these reasoning steps and is responsible for serving all the tasks nodes in this flowchart. The edges represent sequentiality and the post-processing flow of information between the different tasks. A node cannot be run unless all the nodes to it's left, connected to it either directly or indirectly, have been run. For instance, Task 13 can only run when Task 1, 11 and 12 have been run.}
    \label{fig:rcflowchart}
\end{figure*}

\subsection{Rationale Annotation}\label{rationale_annotation}
We annotate a small number of examples from Hotpot-QA dataset which adhere to the step-by-step rationale scheme described in the previous two sub-sections. In recent work on few-shot learning \cite{gao-etal-2021-making}, it has been shown that access to a large development set under a few-shot supervision regime creates an unrealistic few-shot setting. To mimic a realistic few-shot setup we pick the development set to be either smaller than or of the same size as of the training set. In total we annotate 98 bridge and 50 comparison type questions from the training set of Hotpot-QA and 32 bridge and 24 comparison type questions from the validation set.

\subsection{Reasoning Circuit for Multi-hop Question Generation}
The tasks described above are arranged as shown in \autoref{fig:rcflowchart}. It is an acyclic graph where information flows from left to right, with generative or control tasks being performed at each node. At the left entry point, the system gets two passages and a pre-defined answer as input, and after being processed through all the reasoning steps, multihop questions are produced at the other end. Initially, the question type $q_{type}$ is determined from passing the inputs through the Task 1 prompt. After this step, if the question type is determined to be comparison, then Tasks 11, 12 and 13 are run sequentially to generate a comparison multi-hop question. Otherwise, if the question type is bridge, then Tasks 2 through 10 are run to generate a bridge multi-hop question. 

Our proposed few-shot mixture of reasoning tasks framework is implemented by finetuning a pretrained bi-directional encoder-decoder masked language model (MLMs), instead of the auto-regressive decoder-only language models such as GPT-2 \cite{gpt2} and LaMDA \cite{lambda} since MLMs are known to be superior for question answering tasks, and have greatly improved parameter efficiency compared to auto-regressive language models \cite{t0,flan} when trained on mixture of tasks. We use the standard encoder-decoder objective of maximizing the log likelihood of text in the ground truth target. For creating input and output prompts, we closely follow the templates proposed by \citet{chada-natarajan-2021-fewshotqa} where prompts are aligned to the format used during the MLM pretraining.

We know that multi-hop reasoning is a complex natural language task and to build performant systems for this task, prior art infused many basic language skills such as named entity recognition and co-reference resolution into their systems, in the form of external tools as noted in related work section. Also, recent findings show that first fine-tuning models with intermediate tasks like question answering and natural language inference before further finetuning \cite{vu-etal-2020-exploring} them on few-shot examples of target task improves few-shot performance on the target task by priming the model with basic language understanding skills. For these reasons, we choose to implement our framework with the T5 transformer\cite{t5} that has already been pretrained on a variety of downstream tasks in addition to unlabelled text. 

\section{Experiments}
We conduct all our experiments on T5-3b v1.0 model with 3 billion parameters\footnote{https://github.com/google-research/text-to-text-transfer-transformer}. 
As a baseline, we consider tuning the same T5-3b model to directly generate multi-hop questions, without any rationale input and provided with only the two passages and answer as input. For the baseline we use a simple prompt in Appendix \ref{sec:appendix}. For the reasoning circuit experiment we utilise rationale steps prompts described in Appendix \ref{sec:appendix} to generate a mixture of task examples from our few-shot annotation. We use a learning rate of 0.00002 with no warm-up and keep constant learning rate throughout. We train for 35 epochs or 5000 steps whichever is higher. 

For fair comparison, we follow the data splits similar to SGGDQ-DP \cite{pan-etal-2020-semantic}, QA4QG \cite{qa4qg} and SGGDQ-DP \cite{wang-etal-2020-answer} to get 90,440 training examples and 6,072 test examples respectively, note however that instead of using the entire test set as validation set as done in prior work we only validate with same or lower  number of examples as that used for training, see section \ref{rationale_annotation} for details. We use two settings as input to the encoder: 1. The original training data in HotpotQA, in which each question is paired with two long documents, and 2. a pre-processed version of the data where only  supporting sentences required to answer the gold question are kept. We conduct experiments with 8, 16, 32, 64, 128 training examples where 75\% of the examples are drawn from the bridge-type while the remaining 25\% examples are of the comparison type, this is done to mimic the distribution of question types in the original HotpotQA dataset. The number of validation examples is equal to the number of training examples until the 32-shot experiment. After this the number of validation examples gets capped at 32 bridge type questions and 24 comparison type questions. We also conduct an experiment with 148 training examples that constitute all collected annotations. We tune on the average of BLEU1, BLEU2, BLEU3 and BLEU4 \cite{papineni-etal-2002-bleu}, METEOR \cite{lavie-agarwal-2007-meteor}, and ROUGE-L\cite{lin-2004-rouge} scores on our few-shot validation sets.

\begin{table*}[!ht]
\centering
\resizebox{\textwidth}{!}{\begin{tabular}{l|cc|cccccc} 
\hline
\multicolumn{1}{c|}{Models} & \# Training & \# Validation & BLEU-1         & BLEU-2         & BLEU-3         & BLEU-4        & METEOR         & ROUGE-L         \\ 
\hline
\multicolumn{9}{c}{Encoder Input: Supporting Fact Sentences}                                                                                                    \\ 
\hline
ASs2s-a$^*$ \cite{kim2019improving}                    & 90,440      & 6,072         & 37.67          & 23.79          & 17.21          & 12.59         & 17.45          & 33.21           \\
SemQG$^*$ \cite{zhang-bansal-2019-addressing}                      & 90,440      & 6,072         & 39.92          & 26.73          & 18.73          & 14.71         & 19.29          & 35.63           \\
F+R+A  \cite{xie-etal-2020-exploring}                    & 90,440      & 6,072         & 37.97          & -              & -              & 15.41         & 19.61          & 35.12           \\
SGGDQ-DP \cite{pan-etal-2020-semantic}               & 90,440      & 6,072         & 40.55          & 27.21          & 20.13          & 15.53         & 20.15          & 36.94           \\
ADDQG \cite{wang-etal-2020-answer}                & 90,440      & 6,072         & 44.34          & 31.32          & 22.68          & 17.54         & 20.56          & 38.09           \\
QA4QG-Large \cite{qa4qg} & 90,440      & 6,072         & 49.55          & 37.91          & 30.79          & 25.70         & 27.44          & 46.48           \\
\citet{cheng-etal-2021-guiding}               & 57,397      & 6,072         & -              & -              & 21.07          & 15.26         & 19.99          & -               \\ 
\hline
Baseline                    & 8           & 8             & 24.40          & 13.76          & 7.49           & 4.50          & 17.18          & 24.63           \\
Reasoning Circuits          & 8           & 8             & 20.19          & 10.85          & 5.93           & 3.65          & 15.53          & 21.45           \\ 
\hdashline
Baseline                    & 16          & 16            & 24.47          & 14.37          & 8.39           & 5.33          & 17.93          & 25.21           \\
Reasoning Circuits          & 16          & 16            & 22.64          & 12.83          & 7.31           & 4.53          & 17.64          & 22.82           \\ 
\hdashline
Baseline                    & 32          & 32            & 28.27          & 17.63          & 10.84          & 7.03          & 21.00          & 27.77           \\
Reasoning Circuits          & 32          & 32            & 26.09          & 15.93          & 9.56           & 6.11          & 20.39          & 25.67           \\ 
\hdashline
Finetuning                  & 64          & 48            & 28.74          & 18.19          & 11.44          & 7.59          & 21.78          & 28.20           \\
Reasoning Circuits          & 64          & 48            & \textbf{29.60} & \textbf{18.62} & \textbf{11.66} & \textbf{7.62} & \textbf{22.35} & \textbf{28.32}  \\ 
\hdashline
Finetuning                  & 128         & 56            & 31.42          & 20.53          & 13.34          & 8.94          & 24.02          & 30.62           \\
Reasoning Circuits          & 128         & 56            & \textbf{32.77} & \textbf{21.58} & \textbf{14.08} & \textbf{9.50} & \textbf{25.51} & \textbf{31.33}  \\ 
\hline
\multicolumn{9}{c}{Encoder Input: Full Document Context}                                                                                                        \\ 
\hline
MultiQG   \cite{su-etal-2020-multi}                  & 90,440      & 6,072         & 40.15          & 26.71          & 19.73          & 15.2          & 20.51          & 35.30           \\
GATENLL+CT \cite{sachan}                 & 90,440      & 6,072         & -              & -              & -              & 20.02         & 22.40          & 39.49           \\
LowResourceQG   \cite{yu-etal-2020-low}            & 9,000       & 6,072         & -              & -              & -              & 19.07         & 19.16          & 39.41           \\
QA4QG-Base$^*$ \cite{qa4qg}                & 90,440      & 6,072         & 43.72          & 31.54          & 24.47          & 19.68         & 24.55          & 40.44           \\
QA4QG-Large$^*$    \cite{qa4qg}             & 90,440      & 6,072         & 46.45          & 33.83          & 26.35          & 21.21         & 25.53          & 42.44           \\ 
\hline
Finetuning                  & 8           & 8             & 24.17          & 13.46          & 7.38           & 4.46          & 16.79          & 24.71           \\
Reasoning Circuits          & 8           & 8             & 17.76          & 8.91           & 4.56           & 2.66          & 13.81          & 19.74           \\ 
\hdashline
Finetuning                  & 16          & 16            & 25.61          & 15.04          & 8.76           & 5.47          & 18.74          & 25.18           \\
Reasoning Circuits          & 16          & 16            & 21.77          & 12.00          & 6.72           & 4.12          & 16.84          & 22.01           \\ 
\hdashline
Finetuning                  & 32          & 32            & 27.04          & 16.75          & 10.23          & 6.64          & 20.31          & 26.39           \\
Reasoning Circuits          & 32          & 32            & 25.29          & 14.94          & 8.69           & 5.55          & 19.46          & 24.62           \\ 
\hdashline
Finetuning                  & 64          & 48            & 28.06          & 17.52          & 10.89          & 7.14          & 21.11          & 27.42           \\
Reasoning Circuits          & 64          & 48            & 27.92          & 16.92          & 10.21          & 6.51          & 21.01          & 26.90           \\ 
\hdashline
Finetuning                  & 128         & 56            & 28.31          & 18.05          & 11.41          & 7.60          & 22.65          & 28.18           \\
Reasoning Circuits          & 128         & 56            & \textbf{30.67} & \textbf{19.58} & \textbf{12.42} & \textbf{8.25} & \textbf{23.86} & \textbf{29.33}  \\
\hline
\end{tabular}}
\caption{Evaluation results on automatic evaluation metrics for few-shot Reasoning Circuits and fine-tuning baseline experiments with different encoder input settings are reported. We mark in bold where reasoning circuits perform better than our baseline. We also show performance of previous MQG methods on the HotpotQA dataset. Note that most of prior work trained on the entire 90K examples in the HotpotQA dataset with the exception of LowResourceQG \cite{yu-etal-2020-low} trained on 9K and  \citet{cheng-etal-2021-guiding} trained on 57K  examples. $^*$ Results as reported by \citet{qa4qg}. \label{table:results}}
\end{table*}
\begin{table*}[!ht]
\centering
\begin{tabular}{c|cc|ccc|cc|cc} 
\toprule
\multirow{2}{*}{Model} & \multicolumn{2}{c|}{Multi-hop} & \multicolumn{3}{c|}{Well formed} & \multicolumn{2}{c|}{Answerable} & \multicolumn{2}{c}{Answer Matching}  \\ 
\cline{2-10}
                       & Yes  & No                      & Yes  & Acceptable & No           & Yes  & No                       & Yes  & No                            \\ 
\hline
Baseline               & 45\% & 55\%                    & 89\% & 2\%        & 9\%          & 87\% & 13\%                     & 75\% & 25\%                          \\
Reasoning Circuits     & 66\% & 34\%                    & 81\% & 4\%        & 15\%         & 89\% & 11\%                     & 79\% & 21\%                          \\
\bottomrule
\end{tabular}
\caption{Human evaluation results.\label{table:human}}
\end{table*}
\section{Results and Analysis}
\subsection{Automatic Evaluation}

The automatic evaluation metrics used are BLEU1, BLEU2, BLEU3, BLEU4, METEOR, and ROUGE-L, which measure similarity between generations and the target reference questions. 

We report the resuls in Table \ref{table:results}. Reasoning Circuits perform better than the baseline for 64- and 128-shot when entire passages are input to the encoder as well as for 128-shot when only supporting sentences are input to the encoder.

We note a substantially reduced performance gap in the METEOR score between prior state-of-the art models trained with 90k training examples and results of our best few-shot experiments. The METEOR metric has certain synonymy matching and stemming modules, in addition to standard exact word matching which not found in other metrics. From this we infer that reference and generated questions may not exactly match each other however could be closer paraphrases of each other.

Though the performances of baseline and Reasoning Circuits are quite close in terms of automatic metrics, we observe (See models generations in \ref{sec:generations}) through manual inspection that the questions generated by Reasoning Circuits lead to more multi-hop questions being generated whereas baseline generations tend to be single-hop questions instead.  Through automatic evaluation these advantages are not reflected.

\subsection{Human Evaluation}
We follow similar human evaluation criteria as \citet{cheng-etal-2021-guiding}, wherein we randomly sample 150 questions from our 128-shot baseline and Reasoning Circuits experiments of the kind where the input to encoder is only supporting sentences. These examples are manually evaluated by a human annotator across the following four dimensions:

\noindent\textbf{Multi-hop}: To check whether a question can be answered from only reading a single passage or both. The annotation is yes if both passages have to be read or no if reading only a single passage answers the question.

\noindent\textbf{Well-formed}: To check whether a question is semantically correct, annotator is asked to mark a question as either yes, acceptable or no. Acceptable is selected if the question is not grammatically correct, but its meaning is still intelligible.

\noindent\textbf{Answerable:} It checks whether a question is
answerable according to the given context. The
annotation is either yes or no.

\noindent\textbf{Answer Matching}: It checks whether the given answer is the correct answer to the question. The annotation is either yes or no.

Table \ref{table:human} report results from human evaluation. Our proposed approach generated +22\% more multi-hop questions than the baseline which fares poorly on this critical measure. The pre-defined answer is found to be the correct answer to the questions generated with our slightly higher chance than the baseline. However, our approach leads to slightly less well formed generations than the baseline model, typically this stems from our approach failing to find the right common noun. In terms of answerability both approaches score evenly.

\section{Discussion and Future Work}
Through automatic and human evaluations we show that larger language models generate similar questions to reference questions with orders of magnitude less labelled data. The proposed approach also is found to generate a much higher percentages of multi-hop questions than the baseline.

One avenue of future work is in the area of self-training. Self-training, involves generating predictions from a weaker model on unlabelled data and using these predictions as additional training data, where the training set now includes the silver predictions on unlabelled data. Self-training may be detrimental for or not improve overall model performance strongly especially when the task is hard for the weaker model \cite{vu-etal-2020-exploring, vu-etal-2021-strata}. Since prediction errors of the weaker model in the silver annotations further reinforce wrong predictions during self-training. In Reasoning Circuits, silver predictions on unlabelled data for many steps of reasoning can be approximately validated with simple heuristics. For instance, predictions from Task 9 can be verified by checking whether they still contains the answer and bridge spans or not, if they do then these predictions can be deemed unfit for self-training. Filtering prediction errors from the initial trained weaker model on unlabeled data should lead to stronger improvements from self-training compared to vanilla self-training.

\section{Conclusion}
In this paper, we propose Reasoning Circuits, a new framework suited to real-world scenarios where the NLP task at hand requires multiple steps of structured reasoning, with only a limited number of available labelled examples, and a small annotation budget, also only a modest deep learning computational infrastructure/budget is accessible. In this work, we apply this framework to the task of few-shot multi-hop question generation which fits all these criteria. We identify structured multi-step rationales that break down this problem into many discrete reasoning steps. Each step in these rationales is treated as a single "task" within a mixture of similar "tasks". The individual tasks can be categorized into control tasks, which control the flow of information between tasks, and generative tasks, that generate free-form text for successive tasks in the Reasoning Circuit. The framework is relatively easy to implement, since only a single generative model is fine-tuned with a mixture of all reasoning steps; at inference time, the same model can generate all reasoning steps sequentially. We show that fine-tuning with only around 64 to 128 labelled rationale examples with our approach is enough to improve automatic evaluation metrics compared to a baseline trained without rationales on the HotpotQA dataset. More importantly, with human evaluation, we find that this framework can strongly improve the central objective of multi-hop QG, to generate challenging questions which cannot be answered from reading only a single passage.

\section{Limitations}

The proposed Reasoning Circuits framework intends to replace the need for thousands of annotated examples with a strong inductive bias of structured rationales. There is two issues with this approach at a conceptual level: 

\noindent1. It may not always be possible to break down a multi-step reasoning problem cleanly into discrete reasoning steps, and another related issue it increasing complexity of the circuit with the complexity of the task.

\noindent2. For the design of these reasoning circuits a researcher must develop a thorough understanding of this reasoning task, so that the final circuit design broadly covers all possible types of reasoning problems expected to be solved. An under- or ill- designed reasoning circuit may cause the system to either not support a certain portion of problems or produce non-sensical outputs.

Essentially, there is trade off between a tighter control over reasoning by investing in a deep understanding of the problem leading to a comprehensive reasoning circuit design and lower annotations budget, versus, less control over logic and depending on a large number of annotations which allow the model to discover this logic on its own at much higher cost of large scale annotations budget.

At the implementation and operations level one of the the key limitations our proposed system is the number of inference steps to solve the problem. The number of times model inference may be needed to solve a single example is equal the length of the longest task sequence chain in the reasoning circuit. One possible solution for this could be by training the model to solve the entire problem by generating all the steps of reasoning and the target string in a single inference step and could massively reduce inference time and costs.

\bibliography{anthology,custom}
\bibliographystyle{acl_natbib}

\appendix

\section{Task Prompts}
\label{sec:appendix}
We provide the prompts used for training T5 below. $SentinelToken_i$ refers special sentinel tokens used while pretraining T5 model. Common entities were filled in for Tasks 1 and 2 using top three longest contiguous sub-sequences found both in $p_1$ and $p_2$ from which common English words and words that are not capitalised indicating common nouns were removed, we also used Flair NER\footnote{https://github.com/flairNLP/flair} library to generate these entities and kept the ones shared across both $p_1$ and $p_2$.
\\
\textbf{Baseline} Context 1: $p_1$ Context 2: $p_2$ Answer: $a$ Question type: $SentinelToken_0$
\\
\textbf{Task 1} Context 1: $p_1$ Context 2: $p_2$ Answer: $a$ Common entities found: $SentinelToken_0$ Question type: $SentinelToken_1$
\\
\textbf{Task 2} Context 1: $p_1$ Context 2: $p_2$ Answer: $a$ Common entities found: $SentinelToken_0$ Question type: $SentinelToken_1$
\\
\textbf{Task 3} Answer: $a$ is $SentinelToken_0$ in context: $p_i$
\\
\textbf{Task 4} Entities: $a$ and $b$ are $SentinelToken_0$.
\\
\textbf{Task 5} Context: $p_i$ Bridge entity: $b$ Answer: $a$ Assertion: $SentinelToken_0$
\\
\textbf{Task 6} Context: $p_i$ Bridge entity: $b$ Assertion: $SentinelToken_0$
\\
\textbf{Task 7} Bridge entity: $b$ Assertion 1: $s_1$ Assertion 2: $s_2$ Combined: $SentinelToken_0$
\\
\textbf{Task 8} Removing bridge entity: $b$ from: $c$ We get: $SentinelToken_0$
\\
\textbf{Task 9} Contract answer entity $a$ from: $c-b$ We get: $SentinelToken_0$
\\
\textbf{Task 10} Turn: $c-b-a$ into question: $SentinelToken_0$
\\
\textbf{Task 11} Context 1: $p_1$ Context 2: $p_2$ Answer: $a$ Assertion from Context 1: $SentinelToken_0$ Assertion from Context 2: $SentinelToken_1$
\\
\textbf{Task 12} Assertion 1: $s_1$ Assertion 2: $s_2$ Combine, compare and think: $SentinelToken_0$
\\
\textbf{Task 13} Combined assertion: $c$ Answer: $a$ Question: $SentinelToken_0$
The outputs prompt for each of these tasks is to generate the expected output items of each task preceeded by $SentinelToken_0$ and $SentinelToken_1$.

\section{Generated Examples}
We provide actual examples of generations from our baseline model and Reasoning Circuits (128-shot, encoder input: Supporting sentences from the passage only)

\label{sec:generations}
\newcommand*{\myfont}{\fontfamily{lmtt}\selectfont}
{\myfont
{\parindent0pt 
============== Q1 ==================

\textbf{P1}: John Updike - John Hoyer Updike (March 18, 1932  January 27, 2009) was an American novelist, poet, short story writer, art critic, and literary critic.

\textbf{P2}: Bret Easton Ellis - Bret Easton Ellis (born March 7, 1964) is an American author, screenwriter, and short story writer.

\textbf{Gold Question}: What profession was both John Updike and Bret Easton Ellis ?

\textbf{Answer}: short story writer

\textbf{Gold Type}: comparison

---- Generations ----

\textbf{Reasoning Circuits question type}: comparison

\textbf{Reasoning Circuits}: What kind of writers were Bret Easton Ellis and John Updike?

\textbf{Baseline}: What job did both Bret Easton Ellis and John Updike have in common?

============== Q2 ==================

\textbf{P1}: Xingcheng - Xingcheng (), former name Ningyuan (), is a county-level city of southwest Liaoning province, China, with a population of approximately 140,000 urban inhabitants, and is located on the Liaodong Bay, i.e. the northern coast of the Bohai Sea.

\textbf{P2}: Ulan Hot - Ulanhot (Mongolian- ; Cyrillic-   ; Latin transliteration- "Ulaan qota"; ), formerly known as Wangin Sm, alternatively Wang-un Sme, Ulayanqota (Red City) in Classical Mongolian, and Wangyehmiao or Wangyemiao () in Chinese prior to 1947, is a county-level city and the administrative center of Hinggan League in the East of Inner Mongolia autonomous region.

\textbf{Gold Question}: Is Xingcheng or Ulan Hot located in the Inner Mongolia region of China?

\textbf{Answer}: Ulanhot

\textbf{Gold Type}: comparison

---- Generations ----

\textbf{Reasoning Circuits question type}: comparison

\textbf{Reasoning Circuits}: Which city is in the East of Inner Mongolia autonomous region, Ulanhot or Xingcheng?

\textbf{Baseline}: What is the name of this city, which was formerly known as Wangin Sm?

============== Q3 ==================

\textbf{P1}: French Spaniel -  The breed is recognised by Canadian and international kennel clubs but not by The Kennel Club (UK).

\textbf{P2}: Beagle - The Beagle is a breed of small hound, similar in appearance to the much larger foxhound.

\textbf{Gold Question}: Are both French Spaniel and Beagle universally recognized breeds?

\textbf{Answer}: no

\textbf{Gold Type}: comparison

---- Generations ----

\textbf{Reasoning Circuits question type}: comparison

\textbf{Reasoning Circuits}: Are the Beagle and the foxhound similar in appearance?

\textbf{Baseline}: Are the Beagle and the French Spaniel both breeds of small hound?

============== Q4 ==================

\textbf{P1}: The Tempest (1979 film) - The Tempest is a 1979 film adaptation of William Shakespeare's play of the same name. Directed by Derek Jarman, with Heathcote Williams as Prospero, it also stars Toyah Willcox, Jack Birkett and Helen Wellington-Lloyd from Jarman's previous feature, "Jubilee" (1977), as well as his long-time cohort Karl Johnson.

\textbf{P2}: Heathcote Williams - John Henley Heathcote-Williams (15 November 1941  1 July 2017), known as Heathcote Williams, was an English poet, actor, political activist and dramatist. He wrote a number of book-length polemical poems including "Autogeddon", "Falling for a Dolphin" and "Whale Nation", which in 1988 became, according to Philip Hoare, "the most powerful argument for the newly instigated worldwide ban on whaling."

\textbf{Gold Question}: What is the title of the 1979 film adaptation of William Shakespeare's play in which the English poet, actor, political activist and dramatist who wrote wrote a number of book-length polemical poems such as "Autogeddon", "Falling for a Dolphin" and "Whale Nation" played a main character?

\textbf{Answer}: The Tempest

\textbf{Gold Type}: bridge

---- Generations ----

\textbf{Reasoning Circuits question type}: bridge

\textbf{Reasoning Circuits}: What is the film in which an English poet, actor, political activist and dramatist wrote a number of book-length polemical poems including "Whale Nation"?

\textbf{Baseline}: What 1979 film starring Heathcote Williams was directed by Derek Jarman?

============== Q5 ==================

\textbf{P1}: Achel Abbey - The Trappist Abbey of Achel or Saint Benedictus-Abbey or also Achelse Kluis (which means hermitage of Achel), which belongs to the Cistercians of Strict Observance, is located in Achel in the Campine region of the province of Limburg (Flanders, Belgium). The abbey is famous for its spiritual life and its brewery, which is one of few Trappist beer breweries in the world.

\textbf{P2}: Trappist beer -  Eleven monasteries  six in Belgium, two in the Netherlands and one each in Austria, Italy and United States  currently brew beer and sell it as "Authentic Trappist Product".

\textbf{Gold Question}: The Trappist Abbey of Achel produces and sells what as an "Authentic Trappist Product"?

\textbf{Answer}: Trappist beer

\textbf{Gold Type}: bridge

---- Generations ----

\textbf{Reasoning Circuits question type}: bridge

\textbf{Reasoning Circuits}: What kind of beer is brewed at Achel Abbey, which is famous for its spiritual life and its brewery?

\textbf{Baseline}: What is brewed at the Trappist Abbey of Achel?

============== Q6 ==================

\textbf{P1}: Marc Bolan -  He was best known as the lead singer of the glam rock band T. Rex.

\textbf{P2}: Metal Guru - "Metal Guru" is a song by the British rock band T. Rex, written by Marc Bolan. It was the band's fourth (and final) number one on the UK Singles Chart when it topped the chart for four weeks from MayJune 1972.

\textbf{Gold Question}: In the summer of 1972, "Metal Guru" was the last UK number one for T. Rex and its lead singer. What was his name?

\textbf{Answer}: Marc Bolan

\textbf{Gold Type}: bridge

---- Generations ----

\textbf{Reasoning Circuits question type}: bridge

\textbf{Reasoning Circuits}: Who is the best known person who also wrote the song Metal Guru and who also played the guitar for the band T. Rex?

\textbf{Baseline}: Who wrote the song Metal Guru for the glam rock band T. Rex?

============== Q7 ==================

\textbf{P1}: Estonian Hound -  It was bred in 1947 when the Soviet Union's national economy ministry decided that every country in the Union must have its own dog breed.

\textbf{P2}: English Water Spaniel - The English Water Spaniel is a breed of dog that has been extinct since the first part of the 20th century, with the last specimen seen in the 1930s.

\textbf{Gold Question}: Which breed was bred first, the English Water Spaniel or the Estonian Hound?

\textbf{Answer}: The English Water Spaniel

\textbf{Gold Type}: comparison

---- Generations ----

\textbf{Reasoning Circuits question type}: comparison

\textbf{Reasoning Circuits}: Which breed of dog has been extinct since the first part of the 20th century, the English Water Spaniel or the Estonian Hound?

\textbf{Baseline}: Which breed of dog was bred in 1947, the Estonian Hound or the English Water Spaniel?

============== Q8 ==================

\textbf{P1}: Tommy's Honour -  The film is directed by Jason Connery, and the father and son are portrayed by Peter Mullan and Jack Lowden.

\textbf{P2}: Jack Lowden - Jack Andrew Lowden (born 2 June 1990) is a Scottish stage, television, and film actor. Following a highly successful and award-winning four-year stage career, his first major international onscreen success was in the 2016 BBC miniseries "War \& Peace", which led to starring roles in feature films.

\textbf{Gold Question}: Tommy's Honour was a drama film that included the actor who found success with what 2016 BBC miniseries?

\textbf{Answer}: War \& Peace

\textbf{Gold Type}: bridge

---- Generations ----

\textbf{Reasoning Circuits question type}: bridge

\textbf{Reasoning Circuits}: In which 2016 BBC miniseries did a Scottish actor have his first major international onscreen success?

\textbf{Baseline}: What was the first major international onscreen success of this Scottish actor, who played the father in the film Tommys Honour?

============== Q9 ==================

\textbf{P1}: Moho House - "Moho House" is the twenty first episode of the twenty-eighth season of the animated television series "The Simpsons", and the 617th episode of the series overall.

\textbf{P2}: The Simpsons (season 28) -  On May 4, 2015, Fox announced that "The Simpsons" had been renewed for season 28.

\textbf{Gold Question}: How many seasons has a popular tv show had, in which one of the episodes is called Moho House?

\textbf{Answer}: 28

\textbf{Gold Type}: bridge

---- Generations ----

\textbf{Reasoning Circuits question type}: bridge

\textbf{Reasoning Circuits}: Moho House is the twenty first episode of the twenty-eighth season of the animated television series which was renewed for season 28 on May 4, 2015?

\textbf{Baseline}: Moho House is the twenty first episode of which season of the animated television series The Simpsons, and the 617th episode of the series overall?

============== Q10 ==================

\textbf{P1}: Harry S. Truman Supreme Court candidates - During his two terms in office, President Harry S. Truman appointed four members of the Supreme Court of the United States- Chief Justice Fred M. Vinson, Associate Justice Harold Burton, Associate Justice Tom C. Clark, and Associate Justice Sherman Minton.

\textbf{P2}: Fred M. Vinson -  The most prominent member of the Vinson political family, he was the 53rd United States Secretary of the Treasury and the 13th Chief Justice of the United States.

\textbf{Gold Question}: Of four Harry S. Truman Supreme Court candidates, who was the 53rd United States Secretary of the Treasury and the 13th Chief Justice of the United States?

\textbf{Answer}: Fred M. Vinson

\textbf{Gold Type}: bridge

---- Generations ----

\textbf{Reasoning Circuits question type}: bridge

\textbf{Reasoning Circuits}: Who was the most prominent member of the Vinson political family and the 13th Chief Justice of the United States who was appointed by President Harry S. Truman?

\textbf{Baseline}: During his two terms in office, President Harry S. Truman appointed four members of the Supreme Court of the United States, including the most prominent member of the Vinson political family, he was the 53rd United States Secretary of the Treasury and the 13th Chief Justice of the United States.

============== Q11 ==================

\textbf{P1}: Lee Ranaldo - Lee Mark Ranaldo (born February 3, 1956) is an American musician, singer-songwriter, guitarist, writer, visual artist and record producer, best known as a co-founder of the alternative rock band Sonic Youth. In 2004, "Rolling Stone" ranked Ranaldo at number 33 on its "Greatest Guitarists of All Time" list.

\textbf{P2}: Mikael kerfeldt - Lars Mikael kerfeldt (born 17 April 1974) is a Swedish musician, prominently known as the lead vocalist, guitarist, and primary songwriter of progressive death metal band Opeth, as well as being the former vocalist of death metal supergroup Bloodbath. He was also guitarist for the "one-off" band Steel, and is part of the collaboration Storm Corrosion with Steven Wilson.

\textbf{Gold Question}: Mikael kerfeldt and Lee Ranaldo were this kind of instrumentalist in their respective bands. 

\textbf{Answer}: Guitarists

\textbf{Gold Type}: comparison

---- Generations ----

\textbf{Reasoning Circuits question type}: comparison

\textbf{Reasoning Circuits}: What is common between Lee Ranaldo and Mikael Kerfeldt?

\textbf{Baseline}: What do Lee Ranaldo and Mikael   kerfeldt have in common?

============== Q12 ==================

\textbf{P1}: Bent Hamer - Bent Hamer (born 18 December 1956) is a film director, writer and producer, born in Sandefjord, Norway in 1956.

\textbf{P2}: Yasujir Ozu - Yasujir Ozu (  , Ozu Yasujir , 12 December 1903  12 December 1963) was a Japanese film director and screenwriter.

\textbf{Gold Question}: Between Yasujir Ozu and Bent Hamer, who has more creative titles?

\textbf{Answer}: Bent Hamer

\textbf{Gold Type}: comparison

---- Generations ----

\textbf{Reasoning Circuits question type}: bridge

\textbf{Reasoning Circuits}: What film director, writer and producer, born in Sandefjord, Norway in 1956 worked for Yasujiro Ozu who was a Japanese film director and screenwriter?

\textbf{Baseline}: Which film maker was born in Sandefjord, Norway, Bent Hamer or Yasujir   Ozu?

============== Q13 ==================

\textbf{P1}: Ulli Lommel -  Since 1977 he has lived and worked in the USA, where he has written, directed and starred in over 50 movies.

\textbf{P2}: Benjamin Stoloff - Benjamin "Ben" Stoloff (October 6, 1895  September 8, 1960) was an American film director and producer. Stoloff was born in Philadelphia, Pennsylvania. He died in Hollywood, California.

\textbf{Gold Question}: Who has lived longer in the USA, Ulli Lommel or Benjamin Stoloff?

\textbf{Answer}: Stoloff

\textbf{Gold Type}: comparison

---- Generations ----

\textbf{Reasoning Circuits question type}: comparison

\textbf{Reasoning Circuits}: Which film director and producer lived and worked in the USA, Ulli Lommel or Benjamin Stoloff?

\textbf{Baseline}: What American film director and producer born in Philadelphia, Pennsylvania, died in Hollywood, California, Ulli Lommel or Benjamin Stoloff?

============== Q14 ==================

\textbf{P1}: Five Finger Death Punch discography -  After Hook replaced previous guitarist Darrell Roberts, 5FDP released "War Is the Answer" in 2009 which reached the top ten of the "Billboard" 200 and was certified platinum by the RIAA.

\textbf{P2}: War Is the Answer -  It was released on September 22, 2009 through Prospect Park.

\textbf{Gold Question}: Which Five Finger Death Punch discography albulm was released on September 22, 2009 through Prospect Park?

\textbf{Answer}: War Is the Answer

\textbf{Gold Type}: bridge

---- Generations ----

\textbf{Reasoning Circuits question type}: bridge

\textbf{Reasoning Circuits}: What song did Five Finger Death Punch release on September 22, 2009 through Prospect Park?

\textbf{Baseline}: Which album by Five Finger Death Punch was released in 2009, after Hook replaced previous guitarist Darrell Roberts?

============== Q15 ==================

\textbf{P1}: Wolfgang Becker - Wolfgang Becker (born 22 June 1954) is a German film director and writer.

\textbf{P2}: Jacques Audiard - Jacques Audiard (] ; born 30 April 1952) is a French film director and screenwriter.

\textbf{Gold Question}: Are Wolfgang Becker and Jacques Audiard both German film directors?

\textbf{Answer}: no

\textbf{Gold Type}: comparison

---- Generations ----

\textbf{Reasoning Circuits question type}: comparison

\textbf{Reasoning Circuits}: Are Wolfgang Becker and Jacques Audiard both writers?

\textbf{Baseline}: Are Jacques Audiard and Wolfgang Becker both film directors and writers?

============== Q16 ==================

\textbf{P1}: ECAC Hockey -  The conference used to be affiliated with the Eastern College Athletic Conference, a consortium of over 300 colleges in the eastern United States.

\textbf{P2}: Colgate Raiders women's ice hockey - The Colgate Raiders women's ice hockey team is an NCAA Division I ice hockey team that represents Colgate University and play in ECAC Hockey.

\textbf{Gold Question}: What athletic conference did the conference that the Colgate Raiders women's ice hockey team play in used to be affiliated with?

\textbf{Answer}: Eastern College Athletic Conference

\textbf{Gold Type}: bridge

---- Generations ----

\textbf{Reasoning Circuits question type}: bridge

\textbf{Reasoning Circuits}: The Colgate Raiders women's ice hockey team is an NCAA Division I ice hockey team that represents Colgate University and play in a conference which used to be affiliated with what?

\textbf{Baseline}: The conference used to be affiliated with which consortium of over 300 colleges in the eastern United States?

============== Q17 ==================

\textbf{P1}: The Futureheads - The Futureheads were an English post-punk band from Sunderland. consisting of Ross Millard (vocals and guitar), Barry Hyde (vocals and guitar), David "Jaff" Craig (bass guitar) and Dave Hyde (drums).

\textbf{P2}: Marcy Playground - Marcy Playground is an American alternative rock band consisting of three members- John Wozniak (lead vocals, guitar), Dylan Keefe (bass), and Shlomi Lavie (drums).

\textbf{Gold Question}: Which band has more members, The Futureheads or Marcy Playground?

\textbf{Answer}: The Futureheads

\textbf{Gold Type}: comparison

---- Generations ----

\textbf{Reasoning Circuits question type}: comparison

\textbf{Reasoning Circuits}: Which band was an English post-punk band, The Futureheads or Marcy Playground?

\textbf{Baseline}: Which post-punk band had members from both Marcy Playground and The Futureheads?

============== Q18 ==================

\textbf{P1}: General Motors Technical Center - The GM Technical Center is a General Motors facility in Warren, Michigan.

\textbf{P2}: Warren, Michigan - Warren is a city in Macomb County in the U.S. state of Michigan.

\textbf{Gold Question}: In what county is the General Motors Technical Center located?

\textbf{Answer}: Macomb County

\textbf{Gold Type}: bridge

---- Generations ----

\textbf{Reasoning Circuits question type}: bridge

\textbf{Reasoning Circuits}: General Motors Technical Center is in what city in what county?

\textbf{Baseline}: The General Motors Technical Center is a General Motors facility in Warren, Michigan, in which U.S. state?

============== Q19 ==================

\textbf{P1}: Balfour Declaration - The Balfour Declaration was a public statement issued by the British government during World War I announcing support for the establishment of a "national home for the Jewish people" in Palestine, then an Ottoman region with a minority Jewish population.

\textbf{P2}: Declaration to the Seven - The Declaration to the Seven was a document written by the British diplomat Sir Henry McMahon and released on June 16, 1918 in response to a memorandum issued anonymously by seven Syrian notables in Cairo who were members of the newly formed Party of Syrian Unity, established in the wake of the Balfour Declaration and the November 23, 1917 publication by the Bolsheviks of the secret May 1916 Sykes-Picot Agreement between Britain and France.

\textbf{Gold Question}: Party of Syrian Unity was established in the wake of a public statement that announced what ?

\textbf{Answer}: support for the establishment of a "national home for the Jewish people" in Palestine

\textbf{Gold Type}: bridge

---- Generations ----

\textbf{Reasoning Circuits question type}: bridge

\textbf{Reasoning Circuits}: What was the combined assertion of a public statement issued by the British government during World War I announcing support for the establishment of a "national home for the Jewish people" in Palestine and a document written by the British diplomat Sir Henry McMahon and released on June 16, 1918 in response to a memorandum issued anonymously by seven Syrian notables in Cairo who were members of the newly formed Party of Syrian Unity, established in the wake of the Balfour Declaration?

\textbf{Baseline}: The Balfour Declaration was a public statement issued by the British government during World War I announcing what?

============== Q20 ==================

\textbf{P1}: El Paso International Airport - El Paso International Airport (IATA- ELP, ICAO- KELP, FAA LID- ELP) is a public airport four miles (6 km) northeast of downtown El Paso, in El Paso County, Texas, United States.

\textbf{P2}: Grand Forks International Airport - Grand Forks International Airport (IATA- GFK, ICAO- KGFK, FAA LID- GFK) is a public airport five miles (8 km) northwest of Grand Forks, in Grand Forks County, North Dakota.

\textbf{Gold Question}: Which airport Grand Forks International Airport or El Paso International Airport is closer to their town ?

\textbf{Answer}: El Paso International Airport

\textbf{Gold Type}: comparison

---- Generations ----

\textbf{Reasoning Circuits question type}: comparison

\textbf{Reasoning Circuits}: Which airport is located in El Paso County, Texas, United States, Grand Forks International Airport or both?

\textbf{Baseline}: Which airport is farther northeast, El Paso International Airport or Grand Forks International Airport?

============== Q21 ==================

\textbf{P1}: Oklahoma Sooners football - The Oklahoma Sooners football program is a college football team that represents the University of Oklahoma (variously "Oklahoma" or "OU"). The team is currently a member of the Big 12 Conference, which is in Division I Football Bowl Subdivision (formerly Division I-A) of the National Collegiate Athletic Association (NCAA).

\textbf{P2}: Justin Brown (wide receiver) - Justin Brown (born March 10, 1991) is a wide receiver for the Toronto Argonauts of the Canadian Football League (CFL).

\textbf{Gold Question}: A wide receiver for the Toronto Argonauts played college football for a team that represents the University of Oklahoma, which belongs to what conference?

\textbf{Answer}: Big 12 Conference

\textbf{Gold Type}: bridge

---- Generations ----

\textbf{Reasoning Circuits question type}: bridge

\textbf{Reasoning Circuits}: Justin Brown is a wide receiver for the Toronto Argonauts of the Canadian Football League (CFL) that represents the University of Oklahoma, it is a member of which conference?

\textbf{Baseline}: Justin Brown is a wide receiver for the Toronto Argonauts of the Canadian Football League, a team that is a member of which conference?

============== Q22 ==================

\textbf{P1}: D. Napier \& Son - D. Napier \& Son Limited was a British engineering company best known for its luxury motor cars in the Edwardian era and for its aero engines throughout the early to mid-20th century.

\textbf{P2}: Edwardian era - The Edwardian era or Edwardian period of British history covers the brief reign of King Edward VII, 1901 to 1910, and is sometimes extended in both directions to capture long-term trends from the 1890s to the First World War.

\textbf{Gold Question}: Which British engineering company is best known for its luxury motor cars in the era of British history that covers the brief reign of King Edward VII?

\textbf{Answer}: D. Napier \& Son

\textbf{Gold Type}: bridge

---- Generations ----

\textbf{Reasoning Circuits question type}: bridge

\textbf{Reasoning Circuits}: What British engineering company best known for its luxury motor cars in the era which covers the brief reign of King Edward VII?

\textbf{Baseline}: What British engineering company best known for its luxury motor cars in the Edwardian era and for its aero engines throughout the early to mid-20th century?

============== Q23 ==================

\textbf{P1}: Werther - Werther is an opera ("drame lyrique") in four acts by Jules Massenet to a French libretto by douard Blau, Paul Milliet and Georges Hartmann (who used the pseudonym Henri Grmont).

\textbf{P2}: Odyssey - The Odyssey ( ; Greek-  "Odsseia", ] in Classical Attic) is one of two major ancient Greek epic poems attributed to Homer.

\textbf{Gold Question}: Are Werther and The Odyssey both operas?

\textbf{Answer}: no

\textbf{Gold Type}: comparison

---- Generations ----

\textbf{Reasoning Circuits question type}: comparison

\textbf{Reasoning Circuits}: Are Werther and The Odyssey both by Jules Massenet?

\textbf{Baseline}: Are The Odyssey and Werther both written by the same author?

============== Q24 ==================

\textbf{P1}: mile Verdet - Marcel mile Verdet (13 March 1824  3 June 1866) was a French physicist. Verdet did much to champion the early theory of the conservation of energy in France through his editorial supervision of the "Annales de chimie et de physique".

\textbf{P2}: Annales de chimie et de physique - Annales de chimie et de physique (French for "Annals of Chemistry and of Physics") is a scientific journal that was founded in Paris, France, in 1789 under the title "Annales de chimie".

\textbf{Gold Question}: what is the english name of mile Verdets editorial?

\textbf{Answer}: Annals of Chemistry and of Physics

\textbf{Gold Type}: bridge

---- Generations ----

\textbf{Reasoning Circuits question type}: bridge

\textbf{Reasoning Circuits}: Which journal was founded in Paris, France, under the editorial supervision of mile Verdet, who did much to champion the early theory of the conservation of energy in France?

\textbf{Baseline}: What is the name of the scientific journal that was founded in Paris, France, in 1789 under the title Annales de chimie?

============== Q25 ==================

\textbf{P1}: Vinylmation -  Most figures are all shaped with the body of Mickey Mouse but have different themed markings, colors, and patterns.

\textbf{P2}: Mickey Mouse -  He was created by Walt Disney and Ub Iwerks at the Walt Disney Studios in 1928.

\textbf{Gold Question}: The vynil collectible Vinylmation are all shaped with the body of a cartoon character created by who ?

\textbf{Answer}: Walt Disney and Ub Iwerks

\textbf{Gold Type}: bridge

---- Generations ----

\textbf{Reasoning Circuits question type}: bridge

\textbf{Reasoning Circuits}: Vinylmation is shaped with the body of a character which was created by who at the Walt Disney Studios in 1928?

\textbf{Baseline}: Who created the character Mickey Mouse in 1928?
}
}
\end{document}